\def\paperTitle{Python Bindings for a Large C++ Robotics Library:\\ The Case of OMPL}

\def\paperAuthor{\author{
Weihang Guo, Theodoros Tyrovouzis, and Lydia E. Kavraki
\thanks{WG, TT, and LEK are with the Department of Computer Science, Rice University, Houston TX, USA {\tt\small \{wg25, tt88, kavraki\}@rice.edu}. LEK is also affiliated with the Ken Kennedy Institute at Rice University. Work on this paper has been supported in part by NSF 2411219, NSF 2336612, and Rice University funds. 
}}}

\newcommand{\Camera}[1]{{#1}}

\documentclass[letterpaper, 10pt, conference]{ieeeconf}
\IEEEoverridecommandlockouts
\usepackage{amsmath,amssymb,amsfonts}
\usepackage{graphicx}
\usepackage{textcomp}
\usepackage{xcolor}
\usepackage{seqsplit}
\usepackage{etoolbox} %

\newcommand{\textcode}[1]{{\footnotesize\ttfamily\bfseries\seqsplit{#1}}}
\robustify{\textcode} %
\usepackage{amsmath}
\usepackage{algorithm}
\usepackage{algpseudocode}
\usepackage{grffile}
\usepackage{wrapfig,epsfig}
\usepackage{url}
\usepackage{xcolor}
\definecolor{gk}{RGB}{120, 120, 120}
\definecolor{gg}{HTML}{5f9411}
\definecolor{gb}{HTML}{417598}
\definecolor{gr}{HTML}{d15120}
\definecolor{gy}{HTML}{d2ad00}
\usepackage{epstopdf}
\usepackage{bbm}
\usepackage{dsfont}
\usepackage{booktabs}
\usepackage{listings}

\newcommand{\textemph}[1]{{\color{gr}\ttfamily\bfseries #1}}

\newcommand{\mydots}{.\hspace{-.1em}.\hspace{-.1em}.}
\definecolor{refcolor}{RGB}{215, 25, 28}
\usepackage[
    activate   = {true},
    protrusion = false,
    expansion  = true,
    kerning    = true,
    spacing    = true,
    tracking   = false,
    auto       = true,
    selected   = true,
    factor     = 1000,
    stretch    = 50,
    shrink     = 10,
]{microtype}
\usepackage[pdfa,colorlinks,bookmarksopen,bookmarksnumbered,allcolors=refcolor]{hyperref}
\usepackage[
    maxbibnames=99,
    maxcitenames=2,
    natbib=true,
    style=numeric-comp,
    backend=biber,
    sorting=none,
    giveninits=true,
    url=false, 
    doi=false,
    eprint=false,
    isbn=false,
]{biblatex}

\usepackage{subfig}
 
\allowdisplaybreaks

\graphicspath{{./figs/}}

\begin{document}

\ifdefined\isarxiv
    \date{}
    \title{\paperTitle}
    \author{\paperAuthor}
    \begin{titlepage}
    \maketitle
      \begin{abstract}
        Python bindings are a critical bridge between high-performance C++ libraries and the flexibility of Python, enabling rapid prototyping, reproducible experiments, and integration with simulation and learning frameworks in robotics research. Yet, generating bindings for large codebases is a tedious process that creates a heavy burden for a small group of maintainers.
In this work, we investigate the use of Large Language Models (LLMs) to assist in generating nanobind wrappers, with human experts kept in the loop. Our workflow mirrors the structure of the C++ codebase, scaffolds empty wrapper files, and employs LLMs to fill in binding definitions. Experts then review and refine the generated code to ensure correctness, compatibility, and performance.
Through a case study on a large C++ motion planning library, we document common failure modes, including mismanaging shared pointers, overloads, and trampolines, and show how in-context examples and careful prompt design improve reliability. Experiments demonstrate that the resulting bindings achieve runtime performance comparable to legacy solutions. Beyond this case study, our results provide general lessons for applying LLMs to binding generation in large-scale C++ projects.

      \end{abstract}
      \thispagestyle{empty}
    \end{titlepage}
    
    {\hypersetup{linkcolor=black}
    \tableofcontents
    }
    \newpage
\else
    \title{\paperTitle}
    \paperAuthor
    \maketitle
    \begin{abstract}
    Python bindings are a critical bridge between high-performance C++ libraries and the flexibility of Python, enabling rapid prototyping, reproducible experiments, and integration with simulation and learning frameworks in robotics research. Yet, generating bindings for large codebases is a tedious process that creates a heavy burden for a small group of maintainers.
In this work, we investigate the use of Large Language Models (LLMs) to assist in generating nanobind wrappers, with human experts kept in the loop. Our workflow mirrors the structure of the C++ codebase, scaffolds empty wrapper files, and employs LLMs to fill in binding definitions. Experts then review and refine the generated code to ensure correctness, compatibility, and performance.
Through a case study on a large C++ motion planning library, we document common failure modes, including mismanaging shared pointers, overloads, and trampolines, and show how in-context examples and careful prompt design improve reliability. Experiments demonstrate that the resulting bindings achieve runtime performance comparable to legacy solutions. Beyond this case study, our results provide general lessons for applying LLMs to binding generation in large-scale C++ projects.

    \end{abstract}
\fi

\section{Introduction}\label{sec:intro}

\begin{figure*}[!ht]
\vspace{5px}
    \centering
    \includegraphics[width=0.7\linewidth]{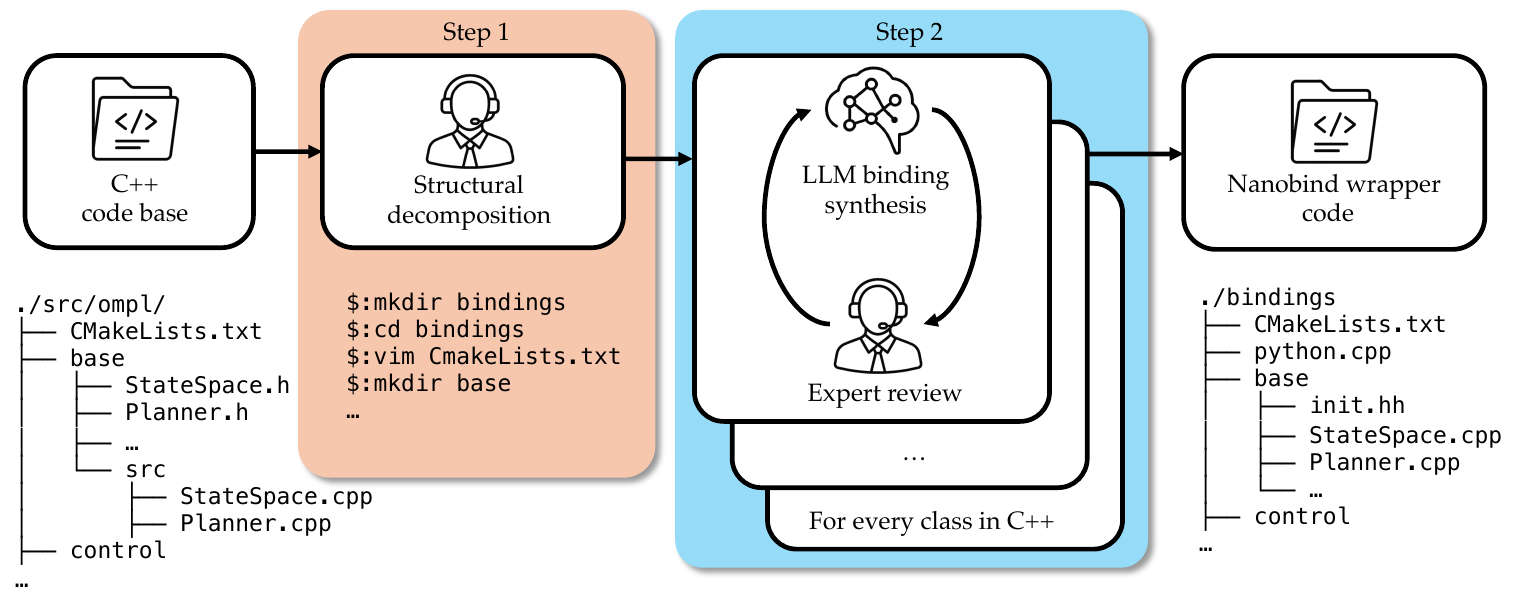}
    \caption{The process begins with the C++ codebase, where human experts analyze the library structure and create scaffolded wrapper files. LLMs are then employed to automatically synthesize the corresponding nanobind binding code. Finally, human experts review, refine, and validate the generated code.}
    \label{fig:fig1}
\end{figure*}

Python has become indispensable in modern robotics and machine learning. Many widely used simulators, such as PyBullet~\cite{pybullet}, Isaac Gym~\cite{isaacgym}, MuJoCo~\cite{mujoco}, are either implemented in Python or provide robust Python APIs for seamless integration. In reinforcement learning~\cite{openaigym}, Python is used for implementing training loops, constructing data pipelines, and managing large-scale experiments. Likewise, recent advances in policy learning~\cite{chi2023diffusion, pi0} rely on Python interfaces to tightly couple learned policies with simulation environments.

A key enabler of this ecosystem is the use of Python bindings, which expose high-performance libraries written in C++ through Python interfaces. These bindings~\cite{pybullet, drake, mujoco} allow researchers to prototype algorithms rapidly and orchestrate experiments in Python while still benefiting from the efficiency of C++-compiled backends. This interoperability is especially important for integrating simulation, optimization, perception, and control modules into unified pipelines. Consequently, well-designed Python bindings are not merely convenient; they form a critical infrastructure for modern robotics research and its deep integration with the broader ecosystem.

Boost.Python~\cite{boostpython}, pybind11~\cite{pybind11}, and nanobind~\cite{nanobind} are all popular tools for creating Python bindings for C++ libraries, but they differ significantly in their design and efficiency. The drawback of Boost.Python is its reliance on the large and complex Boost library. This dependency was necessary to support a wide range of C++ compilers, including older, non-standard ones. A major step forward was pybind11~\cite{pybind11}, which is able to produce very light bindings for C++ code by using features introduced in C++11. After the wide adoption of this standard, Boost.Python's binding infrastructure was able to be significantly simplified.
nanobind, created by the same author of pybind11 focuses on faster compilation, smaller binary sizes, and improved runtime performance. In this work, we adopt nanobind for its performance benefits and modern design. Community adoption is another important factor: as of September 2025, Boost.Python has roughly 500 GitHub stars (repo created in 2000), pybind11 has 17.2k stars (repo created in 2015), and nanobind has 3k stars (repo created in 2022).

The process of generating bindings for large libraries remains a tedious task. While tools like SWIG~\cite{swig} provide multi-language support by automatically generating wrapper code, and tools like Py++~\cite{pyplusplus} were created to automate the process for Boost.Python by allowing developers to specify desired classes and functions. Their adoption has been limited due to compatibility issues (not working on GCC 15.2) and relatively small user communities. Recent advances in large language models (LLMs)~\cite{brown2020language} offer a promising alternative: LLMs can automatically produce much of the nanobind boilerplate code, enabling human experts to focus on validation and refinement. This human-in-the-loop workflow significantly reduces development effort while preserving correctness and consistency.

In this work, we investigate the use of LLMs to generate nanobind-based wrappers with a human expert in the loop. As a case study, we focus on the Open Motion Planning Library~(OMPL)~\cite{ompl}, which implements a wide range of state-of-the-art sampling-based motion planning algorithms and contains over 300 C++ classes across multiple levels of abstraction. OMPL's original bindings, generated with Py++, suffer from compatibility issues and are difficult to maintain. To address this, we develop new bindings using nanobind. Our goal is not to produce one-shot perfect code or automatic error correction, but rather to establish a reliable and maintainable pipeline. We further document common failure modes and propose prompt design strategies that improve robustness. Specifically, we focus on the following objectives:
\begin{enumerate}
    \item \emph{Ease of maintenance}: A clear, modular workspace structure and consistent conventions make the binding easy to understand, extend, and update.
    \item \emph{AI-friendly organization}: A file-per-module structure allows automated tools to generate and modify bindings safely with minimal risk of breaking unrelated components.
    \item \emph{Balancing expert effort and LLM assistance}: Our goal is to minimize human expert involvement while ensuring that the code is correct, safe, and aligned with the designer’s preferences.
    \item \emph{Error analysis of LLMs}: We aim to find which parts of the process (e.g., workspace setup in Sec~\ref{sec:codebase} and binding patterns in Sec.~\ref{sec:pattern}) LLMs make the most mistakes in, and check if these mistakes can be fixed by editing the prompt or giving in-context examples, or if have to rely on human experts.
    \item \emph{OMPL-specific challenges}: We want to keep the API as close as possible to the original while redesigning some parts to make them more intuitive.
\end{enumerate}

\section{Background and Related Work}\label{sec:preli}

Modern robotics research sits at the intersection of high-performance C++ systems and increasingly Python-centric machine learning workflows. As learning-driven methods become integral to robotics, the need for seamless integration between these two ecosystems has grown correspondingly. This section reviews the evolving role of Python in robotics, surveys existing techniques for creating Python bindings to C++ libraries, and situates our work within the broader context of recent advances in LLMs for code generation.

\subsection{Python's Growing Role in Robotics}\label{sec:python}

C++ has long been the dominant language in robotics due to its speed, low-level control, and deterministic performance. Many core libraries, including Drake~\cite{drake}, OMPL~\cite{ompl}, and VAMP~\cite{vamp}, are written in C++ and continue to serve performance-critical roles in motion planning and control.

However, Python has become increasingly important in robotics research, driven by the rise of learning-based methods~\cite{chi2023diffusion, pi0} and the demand for rapid prototyping~\cite{guo2024castl, garrett2020pddlstream}. Simulators such as Mujoco~\cite{mujoco}, PyBullet~\cite{pybullet}, and Isaac Gym~\cite{isaacgym} expose Python APIs or are written natively in Python to support these workflows. Even traditionally C++-centric ecosystems such as ROS~\cite{ros} and MoveIt~\cite{moveit} now provide Python interfaces to support scripting and experimentation. 

\subsection{Python Bindings for C++ Libraries}\label{sec:bindings}

There are several mechanisms for integrating C/C++ code with Python.
Foreign Function Interface libraries such as \textcode{ctypes}~(a standard library built on \textcode{libffi})~\cite{ctypes} and the third-party library \textcode{CFFI}~\cite{cffi} allow Python code to load C shared libraries and call functions from them. While these options allow code to be run from Python without changes to the binary and work across Python implementations, this approach is usually slower since all argument conversions are made at runtime in Python code. Additionally, C++ support is limited as its ABI has not been standardized across platforms and compilers.

CPython, instead, offers a C API~\cite{van2010python} which allows developers to write Python extension modules directly in C/C++, as well as interface with the Python interpreter itself using the definitions from \textcode{Python.h}. The C API is mostly designed to provide a stable, basic interface across Python minor versions and is therefore very lean, low-level, and does not offer higher-level support for C++ features or standard libraries.

At a higher level, binding frameworks such as the Boost.Python~\cite{boostpython}, pybind11~\cite{pybind11}, and nanobind~\cite{nanobind} offer more ergonomic interfaces for exposing C++ classes and functions to Python. Boost.Python was one of the earliest widely adopted solutions, but its dependency on the vast Boost ecosystem introduces significant compile-time overhead. pybind11 and nanobind, both authored by Wenzel Jakob, aim to be lighter and more modern: they eliminate the heavy Boost dependencies while maintaining expressive binding syntax. nanobind further optimizes performance, yielding faster compile times, smaller binaries, and lower runtime overhead than pybind11.

Despite these advances, automated tools for generating bindings remain rare. For example, Py++~\cite{pyplusplus} can auto-generate Boost.Python bindings via parsing the source files, but it sees limited use and suffers from compatibility issues due to the diversity of standards and compilers for C++, as well as the complexity of parsing the C++ language itself. \Camera{In principle, a rule-based procedural generator built on top of an abstract-syntax-trees-based binding generator~\cite{binder} could also be used to produce nanobind bindings. However, due to the need for manual design decisions, framework-specific conventions, and handling of complex C++ constructs, such an approach would require significant customization. Exploring hybrid or fully procedural generation remains another interesting direction not covered in this paper.}

Additionally, the binding process should not always be a 1-1 transformation of an interface. C++ and Python are very different languages with distinct features (e.g., dynamic typing, templates, and memory management), which should be considered during interface design. Manual specification of bindings allows us to rethink and create a more natural and efficient interface that suits Python itself. With Python being garbage collected, the ownership and responsibility of destruction of objects and references are also ambiguous and up to design. While such manual intervention is possible in automatic binding generators to a degree, it is more complicated to do so, with less control than what nanobind provides.

\subsection{LLMs for Coding}\label{sec:llm}

Large Language Models (LLMs) such as Codex~\cite{codex}, AlphaCode~\cite{alphacode}, and CodeGen~\cite{codegen} have demonstrated strong capabilities in generating code, refactoring existing implementations, and assisting with software engineering tasks across a variety of domains. Benchmarks like HumanEval~\cite{codex} and SWE-bench~\cite{swebench} are commonly used to evaluate these models' general coding abilities, focusing on algorithmic challenges and common software tasks. However, none of these benchmarks address the specialized task of generating language bindings such as nanobind~\cite{nanobind}, pybind11~\cite{pybind11}, or Boost.Python~\cite{boostpython} wrappers, which require cross-language understanding and adherence to framework-specific conventions.

Given the absence of benchmarks and the small, specialized field of binding generation, the actual capability of LLMs to produce such code remains largely unexplored. Nevertheless, the remarkable performance of modern LLMs on diverse tasks combined with their powerful in-context learning ability motivates us to investigate this domain. This motivates us to develop a workflow for generating nanobind wrappers and document our findings to guide others in this niche but vital area of tooling.

\subsection{Motivation for Selecting OMPL}\label{sec:why_new_binding}

The Open Motion Planning Library (OMPL)~\cite{ompl} is a C++ library that implements a wide range of sampling-based motion planning algorithms, such as RRT~\cite{kuffner2000rrt}, PRM~\cite{kavraki1996prm}, and KPIECE~\cite{csucan2009kinodynamic}. It focuses on the algorithmic core of motion planning and is designed to be flexible, modular, and extensible. With over 300 classes spanning multiple levels of abstraction, OMPL is widely used in both research and applications. OMPL does not handle environment modeling, collision checking, or visualization directly; instead, it is typically integrated with external tools such as physics engines or robot simulators. Therefore, many researchers prefer to interact with it from Python for rapid prototyping, integration with simulators, and coupling with learning frameworks, making efficient and maintainable Python bindings essential.

The current OMPL Python bindings use Py++~\cite{pyplusplus}, a Boost.Python~\cite{boostpython} generator with limited maintenance and a small user community. It lacks full C++ feature support, requiring custom scripts to filter and manually bind functions, and depends on CastXML/pygccxml, restricting compiler versions and architectures. In the latest GCC version~(15.2), it is no longer working. The significance of OMPL and the limitations of its current Python binding make it an ideal case study for exploring modern Python binding approaches: as a widely used, non-trivial C++ library, developing efficient, maintainable, and feature-complete Python bindings could greatly benefit the robotics community.

\section{Binding Robotics Libraries with Nanobind}\label{sec:model}

In this section, we first describe the manual setup of the binding codebase in Section~\ref{sec:codebase}. We then present common binding patterns used across robotics libraries in Section~\ref{sec:pattern}. Section~\ref{sec:ompl_challenge} highlights the specific challenges that OMPL poses for binding. Finally, in Section~\ref{sec:binding_gen}, we outline our methodology for generating class-to-class bindings with the assistance of LLMs.

\subsection{Manual Setup of the Binding Workspace}\label{sec:codebase}
In this subsection, we describe the method for organizing the codebase for bindings, corresponding to Step 1 in Fig.~\ref{fig:fig1}. We argue that it is generally more reliable for maintainers to manually set up the codebase rather than relying on LLMs to improvise. 

First, repositories vary in structure, and maintainers have different preferences for how bindings should be integrated. LLMs do not capture the high-level philosophy behind these organizational choices. For example, when we ask GitHub Copilot (gpt-o4-mini) to create a workspace for OMPL bindings, it either generates a completely separate workspace with all bindings placed in a single file (\textcode{binding.cpp}), or embeds \textcode{binding.cpp} directly under \textcode{ompl/src}. Though it works, it does not align with the maintainer's preferences. Second, when projects include custom CMake modules, LLMs tend to misconfigure them, often over-correcting and introducing new errors. 

In the case of OMPL, all of its C++ sources live under \textcode{src/ompl}. For example, \textcode{StateSpace} and the geometric \textcode{RRT} are defined in \textcode{src/ompl/base/StateSpace.h} and \textcode{src/ompl/geometric/planners/RRT.h}, respectively. To facilitate easy navigation and maintainability, we mirror this layout under \textcode{bindings/}, placing each wrapper in \textcode{bindings/<Module>/<File>.cpp}, where each file defines exactly one function \textcode{ompl::binding::<Module>::init\_<PathToFile>($\cdot$)}. Here, \textcode{<Module>} is one of \textcode{base}, \textcode{geometric}, \textcode{control}, or \textcode{util}. We create the corresponding headers \textcode{bindings/<Module>/init.hh} to declare these \textcode{init\_<PathToFile>($\cdot$)} functions. Finally, we create \textcode{bindings/python.cpp} to initialize the \textcode{\_ompl} module, create the \textcode{base}, \textcode{geometric}, \textcode{control}, and \textcode{util} submodules, and invoke each component’s \textcode{init} function to register its bindings. This layout lets new contributors quickly locate and update the correct binding code and provides clear context for AI-assisted edits.

To ensure the AI assistant generates bindings in the correct place, we first run a setup script that scaffolds the \textcode{bindings/} directory by creating empty \textcode{bindings/.../*.cpp} files mirroring each C++ source path, parses each path to emit a stub \textcode{init\_<PathToFile>()} function, and inserts the required headers including Nanobind’s includes, the module’s \textcode{init.hh}, and the original C++ header. This is Step 1 in Fig.~\ref{fig:fig1}. An example follows:

\begin{figure}[!ht]
    \centering
    \begin{lstlisting}[language=C++]
 #include <nanobind/nanobind.h>
 #include "ompl/base/SpaceInformation.h"
 #include "init.hh"

 namespace nb = nanobind;

 void ompl::binding::base::init_SpaceInformation(nb::module_& m)
 {
 }
\end{lstlisting}
    \caption{The empty backbone is generated by the manually written script. }
    \label{fig:backbone}
\end{figure}

\subsection{Core Binding Patterns in Robotics Libraries}\label{sec:pattern}
To bind a robotics library, we split the bindings of different classes/functions into the following categories based on binding requirements: \emph{POD~(Plain Old Data) class bindings}, \emph{callback bindings}, and \emph{polymorphic bindings}. In this section, we introduce these categories sequentially and provide OMPL examples to illustrate each case.

\emph{\Camera{Direct} bindings}: bindings that expose C++ classes and methods directly to Python via Nanobind, with no Python-side behavior. This is the most common binding style in OMPL. Examples include \textcode{setLow($\cdot$)}, \textcode{setBounds($\cdot$)}, and the other functions and classes shown in Fig.~\ref{fig:direct_example}.

\emph{Callback bindings}: register selected routines defined in Python as callbacks at runtime, allowing C++ code to invoke Python functions dynamically. As shown in Fig.~\ref{fig:callback_example}, the user-defined arbitrary state validator function \textcode{isStateValid($\cdot$)} falls into this category.

\emph{Polymorphic bindings}: enable Python classes to subclass C++ base classes and override virtual methods, supporting cross-language inheritance and method dispatch. As shown in Fig.~\ref{fig:polymorphic_example}, the user-defined \textcode{RandomWalkPlanner}, a subclass of \textcode{ob.Planner}, falls into this category.

\begin{figure}
    \centering
    \begin{minipage}{\linewidth}
            \centering
\begin{lstlisting}[language=Python]
 # create an SE2 state space
 space = ob.SE2StateSpace()

 # set lower and upper bounds
 bounds = ob.RealVectorBounds(2)
 bounds.setLow(-1)
 bounds.setHigh(1)
 space.setBounds(bounds)
\end{lstlisting}
    \caption{Example illustrating the use of \Camera{direct} bindings in Python.}
    \label{fig:direct_example}
    \end{minipage}

    \begin{minipage}{\linewidth}
            \centering
\begin{lstlisting}[language=Python]
 def isStateValid(state):
    # Some arbitrary condition on the state. 
    return state.getX() < .6
 ...
 ss = og.SimpleSetup(...)
 ss.setStateValidityChecker(isStateValid)
\end{lstlisting}
    \caption{Example illustrating the use of callback bindings in Python.}
    \label{fig:callback_example}
    \end{minipage}
    
    \begin{minipage}{\linewidth}
        \centering
        \begin{lstlisting}[language=Python]
 class RandomWalkPlanner(ob.Planner):
    def __init__(self, si):
        super().__init__(si, "RandomWalkPlanner")
        ...
    def solve(self, ptc):
        ...
    ...
 planner = RandomWalkPlanner(...)
\end{lstlisting}
    \caption{Example illustrating the use of polymorphic bindings in Python.}
    \label{fig:polymorphic_example}
    \end{minipage}
\end{figure}

\subsection{Backward Compatibility}\label{sec:ompl_challenge}
For large, established robotics packages, backward compatibility can be challenging, but it can be addressed as we do with OMPL. In OMPL, we need to balance the need to maintain the existing Python interface with the goal of more closely aligning it with the C++ API. In the old binding, some functions are renamed which means that the C++ and Python interfaces do not match, creating a confusing experience for developers working with both. The new binding preserves both the old Python API names for backward compatibility and the original C++ names, giving developers a consistent experience across both languages. For instance, the C++ method \textcode{printSettings(std::ostream\& out = std::cout)} was previously exposed as \textcode{print(si.settings())}, which is not specified in the OMPL documentation. Our new binding supports both \textcode{si.settings()} (returning a string) and \textcode{si.printSettings()} (printing to console), ensuring compatibility and clarity.

Some functions, particularly those related to state creation and callback registration, are not directly compatible with the old bindings. Our new design improves usability by removing the template \textcode{ScopedState} from \textcode{ompl::base}. Instead, all state objects inherit from \textcode{ompl::base::State}, and a runtime type converter (defined in \textcode{bindings/base/spaces/common.hh}) maps a generic \textcode{State*} to the appropriate \textcode{ScopedState}. Fig.~\ref{fig:syntax_diff} illustrates the difference in creating a start configuration between the old and new bindings.
\begin{figure}
    \centering
\begin{lstlisting}[language=Python]
 space = ob.RealVectorStateSpace(2)
 ... # set the boundary

 ### Old binding ###
 start = ob.State(space)
 start()[0] = -1.
 start()[1] = -1.

 ### New binding ###
 start = space.allocState()
 start[0] = -1.
 start[1] = -1.
\end{lstlisting}
    \caption{Difference between the old and new Python bindings syntax for creating a state.}
    \label{fig:syntax_diff}
\end{figure}

In the old binding, \textcode{start()} is not a state object but a method returning a temporary \textcode{ScopedState}, which can be unintuitive. The new binding simplifies this design and also streamlines callback registration. Previously, Python state validators or propagators had to be explicitly wrapped with \textcode{StateValidityCheckerFn} or \textcode{StatePropagatorFn}. This extra step is no longer necessary. As shown in Fig~\ref{fig:callback_diff}, lines 1–16 are identical in both versions; the only difference appears in lines 18-25.

\begin{figure}
    \centering
\begin{lstlisting}[language=Python]
 from functools import partial
 ...
 def isStateValid(si, state):
     # perform collision checking or check if other constraints are satisfied
     return si.satisfiesBounds(state)
 
 def propagate(start, control, duration, state):
     state.setX(start.getX() + control[0] * duration * cos(start.getYaw()))
     ...
     
 # construct the state space we are planning in
 space = ob.SE2StateSpace()
 cspace = oc.RealVectorControlSpace(space, 2)
 ...
 # define a simple setup class
 ss = oc.SimpleSetup(cspace)
     
 ### Old binding ### 
 ss.setStateValidityChecker(ob.StateValidityCheckerFn(partial(isStateValid, ss.getSpaceInformation())))
 ss.setStatePropagator(oc.StatePropagatorFn(propagate))

 ### New binding ###
 ss.setStateValidityChecker(
     partial(isStateValid, ss.getSpaceInformation()))
 ss.setStatePropagator(propagate)

\end{lstlisting}
\caption{Difference between the old and new Python bindings for callback registration.}

    \label{fig:callback_diff}
\end{figure}

\subsection{Binding Generation Using LLM and Expert Review}\label{sec:binding_gen}

Following the manual scaffolding of the binding backbone in Section~\ref{sec:codebase}, we leverage a large language model to generate each Nanobind wrapper as shown in Fig.~\ref{fig:fig1}. For every C++ header under \textcode{src/ompl/}, we provide the model with (1) the header’s contents, (2) its corresponding scaffold in \textcode{bindings/} (see Fig.~\ref{fig:backbone}), and (3) a standardized prompt. Headers defining multiple classes are handled via separate prompts, so one class is bound each time. The LLM’s output is then reviewed and compiled by a human expert.

\textemph{Remark.}This paper does not aim to showcase one-shot perfect code generation or automated error correction. Instead, we present the lessons learned when using LLMs for Nanobind binding: the failure modes we encountered and the prompt-engineering strategies that improved reliability.

\section{Implementation Details and Experiments}\label{sec:experiments}

In this section, we first describe how direct, callback, and polymorphic bindings are implemented in Section~\ref{sec:nanobind_implementation}. Then, in Section~\ref{sec:setup}, we present benchmarks comparing the legacy OMPL bindings with the newly generated nanobind bindings.

\subsection{nanobind Implementation Details}\label{sec:nanobind_implementation}
In this section, we provide implementation details on how direct, callback, and polymorphic bindings can be realized, using OMPL as a case study.

We start with the direct bindings taking \textcode{RealVectorStateSpace} in OMPL as an example, as shown in Fig~\ref{fig:bind_direct}. The snippet below shows how to bind a C++ class with nanobind. \textcode{RealVectorStateSpace} extends \textcode{StateSpace}, so we pass both to \textcode{nb::class\_}. We include \textcode{shared\_ptr.h} to enable automatic management of shared pointers. The constructor and member functions are bound using \textcode{.def()}. Overloaded functions like \textcode{setBounds} require \textcode{nb::overload\_cast}, and lambdas wrap functions like \textcode{printState} to redirect output.

\begin{figure}
    \centering
\begin{lstlisting}[language=C++]
 #include <nanobind/nanobind.h>
 #include <nanobind/stl/shared_ptr.h>
 ...
 #include "ompl/base/spaces/RealVectorStateSpace.h"
 #include "../init.hh"
 
 namespace nb = nanobind;
 namespace ob = ompl::base;
 void ompl::binding::base::initSpaces_RealVectorStateSpace(nb::module_ &m){
     nb::class_<ob::RealVectorStateSpace, ob::StateSpace>(m, "RealVectorStateSpace")
     .def(nb::init<unsigned int>())
     .def("setup", &ob::RealVectorStateSpace::setup)
     .def("setBounds", nb::overload_cast<const ob::RealVectorBounds &>(&ob::RealVectorStateSpace::setBounds))
     .def("printState", [](const ob::RealVectorStateSpace &space, const ob::State *state){ space.printState(state, std::cout); })
     ...;
 }
\end{lstlisting}
\caption{Direct bindings use nanobind.}

    \label{fig:bind_direct}
\end{figure}

We then show how to bind a callback function using \textcode{SpaceInformation.setStateValidityChecker($\cdot$)} as an example. To support callback functions, we include \textcode{<nanobind/stl/function.h>} and bind the method that accepts a \textcode{std::function}. This enables Python users to pass in a custom state-validity-checking function, which will be called from C++ during planning. The \textcode{nb::overload\_cast} ensures the correct overload is registered.

\begin{figure}
    \centering
    \begin{lstlisting}[language=C++]
 #include <nanobind/nanobind.h>
 #include <nanobind/stl/function.h>
 ...
 nb::class_<ompl::base::SpaceInformation>(m, "SpaceInformation")
 .def("setStateValidityChecker", nb::overload_cast<const std::function<bool(const ob::State*)>&>(&ob::SpaceInformation::setStateValidityChecker))
 ...
\end{lstlisting}
    \caption{Callback bindings use nanobind.}
    \label{fig:placeholder}
\end{figure}

To support subclassing and virtual method overrides in Python, we define a trampoline class \textcode{PyPlanner} inheriting from \textcode{ompl::base::Planner}. The macro \textcode{NB\_TRAMPOLINE(ob::Planner, 8)} declares space for 8 virtual overrides. The pure virtual method \textcode{solve($\cdot$)} is bound using \textcode{NB\_OVERRIDE\_PURE}, while the remaining virtual functions use \textcode{NB\_OVERRIDE} for optional Python overrides. Finally, we register the class using \textcode{nb::class\_<ob::Planner, PyPlanner>}, enabling Python users to extend and implement custom planners. The example is shown in Fig.~\ref{fig:trampoline}. 

\begin{figure}[!ht]
    \centering
    \begin{lstlisting}[language=C++]
 #include <nanobind/nanobind.h>
 #include <nanobind/trampoline.h>
 #include "ompl/base/Planner.h"
 #include "init.hh"
 ...
 namespace nb = nanobind;
 namespace ob = ompl::base;

 struct PyPlanner : ob::Planner {
    NB_TRAMPOLINE(ob::Planner, 8); // 8 indicates the number of virtual functions to override
    ob::PlannerStatus solve(const ob::PlannerTerminationCondition &ptc) override {
        NB_OVERRIDE_PURE(solve, ptc);
    }
    ... //other 7 virtual functions
 };

 // bind the Planner class with the trampoline to allow subclassing in Python
 nb::class_<ob::Planner, PyPlanner>(m, "Planner")
    ...;
\end{lstlisting}
    \caption{Trampoline binding for \textcode{ob::Planner} using nanobind. 
    The trampoline class \textcode{PyPlanner} overrides virtual methods with \textcode{NB\_TRAMPOLINE}.}
    \label{fig:trampoline}
\end{figure}

\subsection{Experimental Setups and Benchmarks}\label{sec:setup}

\begin{table*}[!ht]
\vspace{5px}
\centering
\begin{tabular}{lcc}
\toprule
Experiment (wall-clock time in milliseconds) & nanobind& Boost.Python\\
\midrule
Sample in Pybullet & 16.5 $\pm$ 1.6 & 17.5 $\pm$ 1.8 \\
RRT-Connect Python (2D) & 4.4 $\pm$ 1.6 & 4.5 $\pm$ 1.6 \\
RRT-Connect Python (PyBullet) & 60.4 $\pm$ 7.0 & 63.8 $\pm$ 5.5 \\
Constrained Planning Sphere & 1388.0 $\pm$ 650.4 & 1313.1 $\pm$ 586.8 \\
KPIECE Planning & 2346.3 $\pm$ 1993.2 & 3324.5 $\pm$ 2099.6 \\
\bottomrule
\end{tabular}
\caption{Reported results correspond to wall-clock time (in milliseconds), expressed as $\mathrm{mean} \pm \mathrm{std}$. over 500 independent runs. For RRT-Connect in Python 2D and PyBullet, we applied the Interquartile Range method to remove outliers before computing the mean and standard deviation. We show that nanobind (the newer binding) is generally not slower than Boost.Python (the older binding). Together with its stronger community support, better compatibility, and easier maintenance, the proposed pipeline is the better choice for adoption. 
}
\label{tab:result}
\end{table*}
The experiments were conducted on a workstation running Ubuntu 20.04.5 LTS (x86-64) with Linux kernel 6.8.0-64-generic. The software stack includes CMake 4.0.2, Python 3.12.10, and nanobind 2.7.0~\cite{nanobind}. The OMPL~1.7.0~\cite{ompl} with constrained planning capabilities~\cite{kmk19} was compiled with g++ 11.4.0 under the -O3 release configuration. Experiments were conducted on a system with an AMD Ryzen 5 5600X (Zen 3) 6-core/12-thread processor, running at 3.7GHz base frequency with boost up to 4.6GHz. 
The system is equipped with 32GB DDR4 memory (2×16GB) at 3200~MT/s.

To systematically assess the performance impact of cross-language integration in OMPL, we evaluate standard, callback, and polymorphic bindings (explained in  Sec.~\ref{sec:pattern}) between C++ and Python. Specifically, we choose the following tests. 

\begin{figure}[!ht]
\centering
    \subfloat[Constrained planning sphere.]{
        \includegraphics[width=0.3\linewidth]{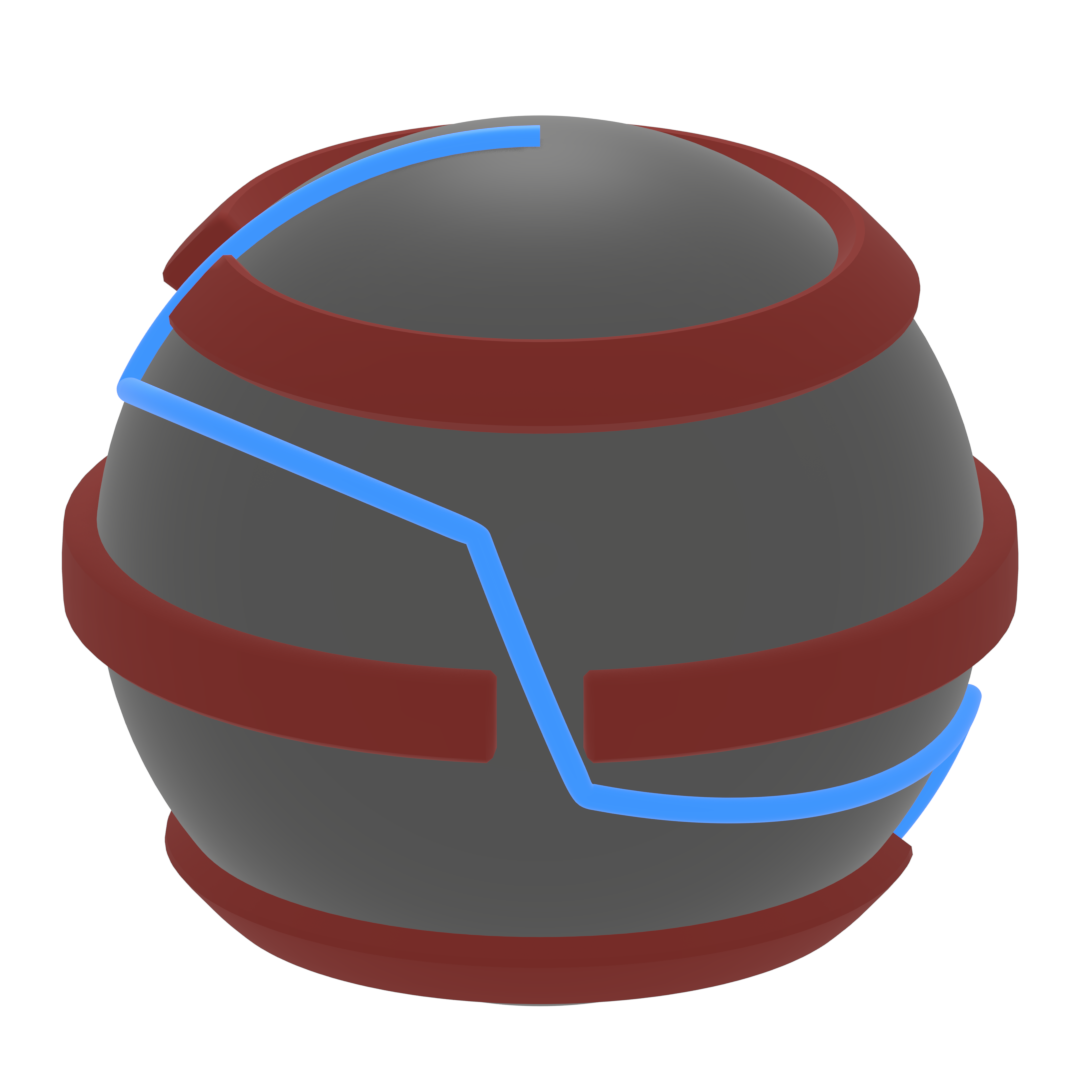}
        \label{fig:sphere}
    }
    \subfloat[Generate robot samples and perform RRT-Connect planning in PyBullet.]{
        \includegraphics[width=0.3\linewidth]{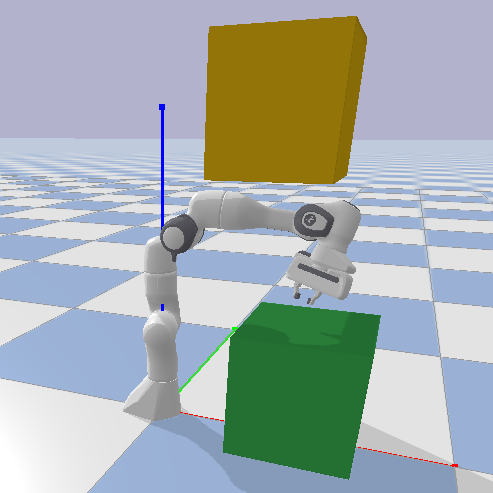}
        \label{fig:sample}
    }
    \subfloat[2D RRT-Connect planning.]{
        \includegraphics[width=0.3\linewidth]{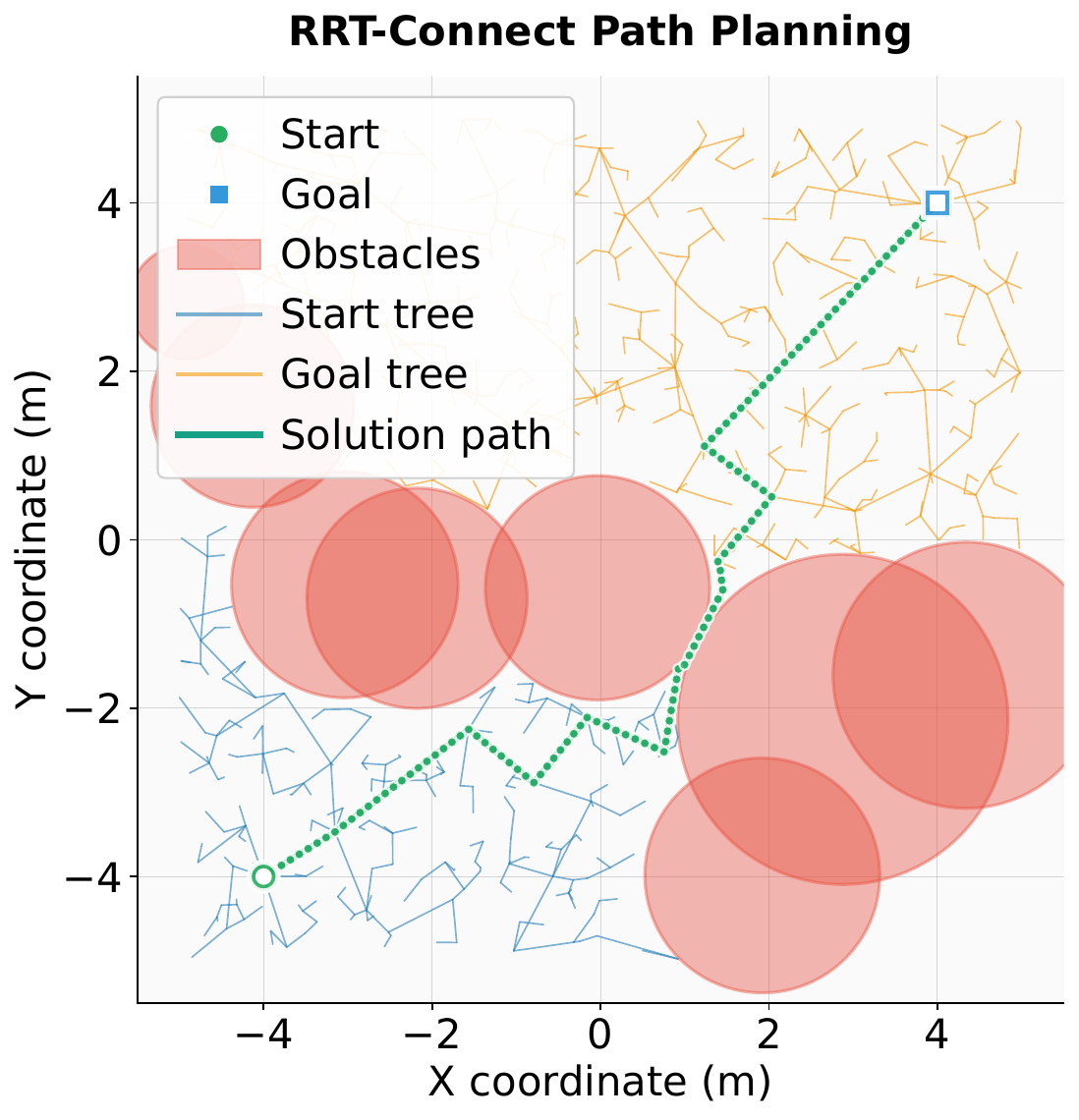}\label{fig:plan2d}
    }
    \caption{Experimental Setup.}
    \label{fig:experiment}
\end{figure}

\emph{Sample in PyBullet}: Sampling was performed in PyBullet~(v3.2.7) using a 7‑DOF Franka Emika Panda arm positioned above a flat plane and two cube obstacles. A screenshot of the setup is shown in Fig~\ref{fig:sample}. The \textcode{StateValidityCheckerFn} was bound to a Python‑defined collision‑checking function. A uniform sampler in the robot’s configuration space was used, and configurations were deemed valid if the manipulator did not collide with the plane, any obstacles, or itself. In each run, 100 collision‑free samples were generated.

\emph{RRT-Connect Python (2D)}: We measure the end-to-end planning time of a Python implementation of RRT-Connect, built by extending the \textcode{ob::Planner} interface. Each experiment is run 500 times, with 8 randomly placed circular obstacles per run. The start (green circle) and goal (blue square) positions, as shown in Fig.~\ref{fig:plan2d}, are randomly sampled, and the timeout was set to 1 second.

\emph{RRT-Connect Python (PyBullet)}: We measure the end-to-end planning time of a Python implementation of RRT-Connect, built by extending the \textcode{ob::Planner} interface. Each experiment is run 500 times, with fixed obstacles shown in Fig.~\ref{fig:sample}. The start and goal configurations are randomly sampled, and the timeout was set to 1 second.

\emph{Constrained Planning Sphere}: The ``sphere" environment is a two-dimensional surface in $\mathbb{R}^3$ defined by the equation $F(q) = \|q\| - 1 = 0.$
The planner must negotiate three longitudinal barriers, each containing a tight passage, to travel from the south pole to the north pole as shown in Fig~\ref{fig:sphere}. The planner is projection-based \textcode{RRTConnect} with 5 5-second timeout. 

\emph{Rigid Body Planning with Control (KPIECE Planning)}: This experiment measures the end‐to‐end planning time for a simple rigid‐body control problem. We construct an $SE(2)$ state space with bounded x–y coordinates ($–1$ to $1$) and a two‐dimensional control space (linear and angular velocity, each in $[–0.3, 0.3]$). A custom Python callback tests state validity against the workspace bounds, and a user‐supplied propagator advances the vehicle by applying controls over a fixed duration. The planner is \textcode{ompl::control::KPIECE1} with a 5-second timeout. 

In our experiment, the nanobind and Boost.Python versions of the state validity checking and propagation functions for KPIECE planning, as well as the collision checking and projection function for constrained sphere planning, are written in Python. This leads to slower wall-clock times compared to native C++ implementations. We intentionally adopt this setup to better reflect the typical usage of Python bindings in practice.
\Camera{For reference, the fully C++ implementation of the constrained planning sphere and KPIECE planning (the last two experiments in Table~\ref{tab:result}) requires $49.9 \pm 16.8~\textrm{ms}$ and $322.3 \pm 424.2~\textrm{ms}$, respectively.}

\section{Discussion on LLM Lessons Learned}\label{sec:discussion}

In this section, we share the lessons learned from using LLMs to generate bindings. For each issue, we discuss whether it can be addressed by refining the prompt, providing in-context examples, or whether it is better resolved by relying on human expertise.

\emph{One-shot for simple static bindings.} When the target class or function has no overloads, no smart pointers (e.g., \textcode{std::shared\_ptr}), and no complex parameter types (e.g., \textcode{std::vector<\ldots>} or \textcode{Eigen} matrices), \textcode{gpt-o4-mini} reliably generates a compilable nanobind stub on the first attempt. Typical successes include constructors, trivial getters/setters, and single-signature methods with POD arguments. Manual edits are usually limited to minor includes or style fixes.

\emph{Errors with \textcode{std::shared\_ptr}.} We found that LLMs often mishandle shared pointers. As shown in Fig.~\ref{fig:failure_ptr}, the most common failure is importing the wrong header (e.g., \textcode{<nanobind/make\_shared.h>} which does not exist) and emitting a pybind11-style holder in \textcode{nb::class\_}. In nanobind, shared-pointer support is enabled via \textcode{<nanobind/stl/shared\_ptr.h>} and \emph{do not} specify a holder type in \textcode{nb::class\_}. Despite explicit prompts, the model sometimes still inserts \textcode{std::shared\_ptr<...>} in the class template, likely conflating pybind11 and nanobind idioms.
\begin{figure}
    \centering
\begin{lstlisting}[language=C++]
 #include <nanobind/make_shared.h> // INCORRECT
 // INCORRECT (pybind11-style specification)
 nb::class_<ob::RealVectorStateSpace,std::shared_ptr<ob::RealVectorStateSpace>>(m, "RealVectorStateSpace");

 // CORRECT (nanobind)
 #include <nanobind/stl/shared_ptr.h>
 nb::class_<ob::RealVectorStateSpace>(m, "RealVectorStateSpace");
 ...
\end{lstlisting}
\caption{Failure case of LLM-generated binding code for shared pointers.}
\label{fig:failure_ptr}
\end{figure}

\emph{Overload handling.} To bind overloaded methods, use \textcode{nb::overload\_cast<Args...>(\&Class::method)} (adding \textcode{nb::const\_} for \textcode{const} overloads). Without explicit guidance, the LLM omits overload disambiguation. After we add this rule to the prompt, the LLM begins inserting \textcode{overload\_cast} even when no overload exists 
which creates false positives. We recommend handling overloads manually to avoid spurious uses of \textcode{nb::overload\_cast}.

\emph{Complex datastructures.} Certain function signatures involve parameter types that nanobind (as of version~2.7.0) cannot handle directly, such as non-const lvalue references as shown in Fig.~\ref{fig:failure_complex}.
\begin{figure}
    \centering
\begin{lstlisting}[language=C++]
 bool checkMotion (const State *s1, const State *s2, std::pair< State *, double > &lastValid) const override
\end{lstlisting}
\caption{Example of a function signature that cannot be directly bound due to unsupported parameter types.}
\label{fig:failure_complex}
\end{figure}

In such cases, LLMs cannot reliably infer which parameter types nanobind supports. We recommend manually filtering out unsupported functions before passing the file to the LLM for binding generation.

\emph{In-context examples for trampolines.}  
Without relevant in-context examples, the LLM fails to correctly generate a trampoline for \textcode{ob::Planner} in all 5 trials.  
Providing only nanobind's trampoline class documentation does not improve results, as all generated implementations remained incorrect.  
In contrast, when a manually written binding for \textcode{ob::Planner} is supplied as an in-context example, the LLM successfully generates correct bindings for \textcode{ob::Goal} and \textcode{ob::Constraint} in 5/5 and 4/5 trials, respectively.

\section{Conclusion}\label{sec:conclusion}

This work demonstrated how nanobind, combined with LLMs and human-in-the-loop review, can streamline the creation of Python bindings for large robotics C++ libraries. Through a structured workflow, we show how to generate bindings that are maintainable, backward compatible, and performant. Using OMPL as a case study, we highlighted lessons learned on common LLM failure modes and effective prompt design, showing how expert oversight ensures correctness and reliability. Our nanobind-based bindings match or exceed the runtime efficiency of the legacy Boost.Python version while overcoming long-standing usability and maintenance issues. These findings point toward a generalizable, LLM-assisted approach for cross-language integration in robotics and beyond.

\section*{Acknowledgements}

\Camera{The authors would like to thank Dr. Mark Moll for reviewing and providing feedback on the new Python bindings and the paper. They also thank Dr. Zachary Kingston (Purdue University, West Lafayette, IN) for discussions that led to the decision to replace the Boost.Python-based bindings with nanobind. The authors are grateful to Emiliano Flores (Rice University, Houston, TX) for benchmarking nanobind under different configurations. Finally, the authors thank the reviewers for their valuable comments, which helped improve the readability of the paper. }

\printbibliography{}

\end{document}